\documentclass[pdflatex,sn-mathphys-num]{sn-jnl}% Math and Physical Sciences Numbered Reference Style
\usepackage{graphicx}%
\usepackage{multirow}%
\usepackage{amsmath,amssymb,amsfonts}%
\usepackage{amsthm}%
\usepackage{mathrsfs}%
\usepackage[title]{appendix}%
\usepackage{xcolor}%
\usepackage{textcomp}%
\usepackage{manyfoot}%
\usepackage{booktabs}%
\usepackage{algorithm}%
\usepackage{algorithmicx}%
\usepackage{algpseudocode}%
\usepackage{listings}%
\theoremstyle{thmstyleone}%
\theoremstyle{thmstyletwo}%

\theoremstyle{thmstylethree}%

\begin{document}

\title[Vision--Language Model--Based Commander]{Tactical Decision for Multi--UGV Confrontation with a Vision--Language Model--Based Commander}

\author[1]{\fnm{Li} \sur{Wang}}\email{liwang11@bit.edu.cn}

\author[2]{\fnm{Qizhen} \sur{Wu}}\email{wuqzh7@buaa.edu.cn}

\author*[3]{\fnm{Lei} \sur{Chen}}\email{bit\_chen@bit.edu.cn}

\affil[1]{\orgdiv{Yangtze Delta Region Academy}, \orgname{Beijing Institute of Technology}, \orgaddress{\city{Jiaxing}, \postcode{314001}, \country{China}}}

\affil[2]{\orgdiv{School of Automation}, \orgname{Beihang University}, \orgaddress{\city{Beijing}, \postcode{100191}, \country{China}}}

\affil*[3]{\orgdiv{Advanced Research Institute of Multidisciplinary Sciences}, \orgname{Beijing Institute of Technology}, \orgaddress{\city{Beijing}, \postcode{100081}, \country{China}}}

\abstract{In multiple unmanned ground vehicle confrontations, autonomously evolving multi--agent tactical decisions from situational awareness remain a significant challenge. Traditional handcraft rule--based methods become vulnerable in the complicated and transient battlefield environment, and current reinforcement learning methods mainly focus on action manipulation instead of strategic decisions due to lack of interpretability. Here, we propose a vision--language model--based commander to address the issue of intelligent perception--to--decision reasoning in autonomous confrontations. Our method integrates a vision language model for scene understanding and a lightweight large language model for strategic reasoning, achieving unified perception and decision within a shared semantic space, with strong adaptability and interpretability. Unlike rule--based search and reinforcement learning methods, the combination of the two modules establishes a full--chain process, reflecting the cognitive process of human commanders. Simulation and ablation experiments validate that the proposed approach achieves a win rate of over 80\% compared with baseline models.}




\maketitle

\section{Introduction}\label{sec1}

In autonomous confrontations involving multiple unmanned ground vehicles~(UGVs), the main responsibility of a tactical commander is to develop coordinated tactical plans for the team based on an in--depth understanding of the battlefield. A significant challenge lies in extracting strategic insights from noisy perceptual data to guide tactical decisions. This challenge is vividly illustrated in the tactical game Frozen Synapse, where players' success depends entirely on their ability to assess the overall situation during the planning phase and make predictive plans. Consequently, the focus is on developing commander intelligence with tactical reasoning and decision--making capabilities in autonomous confrontations.

Attempts to develop such an artificial commander have followed several paradigms in the past. Traditional rule--based approaches offer transparent decision making, but suffer from a critical limitation in their lack of adaptability. Faced with changing tactics, the solution is to continuously expand an already vast rule sets, leading to an explosion in complexity~\cite{hou2023ucav_decision}. In contrast, end--to--end reinforcement learning models offer better adaptability but have inherent unexplainability issues, since their policies are embedded in opaque networks. Furthermore, these models tend to learn policies that map perception directly to actions, which lack strategic--level thinking~\cite{liu2024game_drones,haarnoja2024soccer_rl,qu2023usv_drl}. Even hierarchical reinforcement learning methods, which attempt to combine planning and control, typically have strict predefined interfaces between layers. This makes it difficult to generate novel variations of strategies~\cite{wu2024hierarchical,nian2024large}. Consequently, these limitations motivate the continued pursuit of an intelligent commander that possesses strong adaptability, interpretability, and strategic reasoning.

A growing body of research suggests that natural language is emerging as an effective medium for the development of autonomous reasoning in unmanned systems.  Successes such as the multi--robot task planning framework ReplanVLM~\cite{mei2024replanvlm} and the multi--role soccer robot system LLCoach~\cite{brienza2024llcoach} demonstrate that semantic cues can bridge the conceptual gap between abstract planning and real--world environments, thereby enabling systems to plan tasks and generate strategies autonomously.
The concept can be further extended to adversarial environments with higher challenge, where a macro--level understanding of the situation is essential. Leveraging large language models with extensive world knowledge and powerful reasoning ability enables the realization of an intelligent commander capable of strategic planning and coordination.
In this work, we propose the vision--language model--based commander, which reconstructs the process from perception to decision as a language--based cognitive process. 
Specifically, the cross--modal vision language model~(VLM) acts as the perception front--end, translating situational information into semantic representations, and the lightweight large language model~(LLM) acts as the strategic core for decision and planning. The perception module and the decision module are combined through a shared semantic space to reflect the cognitive and reasoning processes of a human commander.
In addition, we introduce an expert system module to achieve semantic alignment between the VLM and the LLM, and assist our integrated model in adapting to multi--UGV confrontations.

Our main contributions are as follows. 
\begin{itemize}
    \item We model the tactical reasoning process as a natural language--based cognitive process, unifying perception and decision within semantic space to simulate the strategic thinking process of human commanders.
    \item We propose the vision--language model--based commander, which combines a VLM and a lightweight LLM to achieve full--chain integration from perception to tactical decision.
    \item We design an expert system to achieve semantic alignment, as well as model training, and simulation experiments demonstrate the potential of the fine--tuned commander for tactical decision.
\end{itemize}

The rest of this paper is organized as follows.
Section~\ref{sec:related} provides a brief overview of the related work.
Section~\ref{sec:method} presents the proposed methodology and the design of the architecture.
Section~\ref{sec:experients} reports the experimental results, including comparative evaluations and ablation studies.
Finally, Section~\ref{sec:conclusions} concludes the paper and discusses potential directions for future research.

\section{Related Work}\label{sec:related}

In recent years, the capabilities of LLMs have expanded rapidly, evolving from passive text generators to autonomous agents. This transformation drives the application of artificial intelligence in both virtual environments and real--world unmanned systems~\cite{tian2025uavs}.

\subsection{LLMs as Decision--Making Agents}
Early work in board games~\cite{fair2022diplomacy} demonstrated that combining language models with strategic reasoning can reach human levels in the game of Diplomacy.
Complex games filled with negotiations, cooperation, and deception have inspired researchers to explore the ability of LLMs to reason and mimic human interactions. These include iterated Prisoner’s Dilemma~\cite{fontana2025prisoner}, multi--round Werewolf~\cite{jin2024werewolf}, the hidden--role game Avalo~\cite{wang2023avalon} and other interactive games~\cite{bottega2023jubileo}. 

VLMs extend this paradigm by basing linguistic reasoning on visual inputs.
RL4VLM~\cite{zhai2024vlm_rl} prompts VLM to generate chain--of--thought (CoT) reasoning and executable operations for agents based on observations. And in the GameVLM system~\cite{mei2024gamevlm}, multiple decision--making agents are used for robot task planning, and expert agents are introduced to evaluate these plans. At the same time, the CoT mechanism improves the ability of the large language model to gradually reasoning and task decomposition~\cite{10.1145/3719341,mu2023embodiedgpt}. Inspired by these works, we use VLM to observe the overall battlefield environment and introduce an expert system to guide LLM in planning rather than evaluation, thus stimulating the decision--making potential of LLM.

\subsection{LLMs for Unmanned System}
Strategic planning tasks in unmanned systems require a higher level of intelligence~\cite{skaltsis2023review}. A key trend is to reposition LLM as a central inference engine capable of understanding advanced tasks, where LLM can demonstrate its contextual understanding and logical reasoning capabilities. These methods do not rely on specific processes but instead leverage embedded reasoning and world--knowledge within LLM to generate flexible planning in zero--shot and few--shot environments~\cite{liu2024zvqaf,10.1145/3672456}.

Prompt strategies that embed role semantics in the input enable the model to dynamically adjust formations and roles in adversarial environments such as robot football~\cite{brienza2024llcoach}. Crucially, LLMs do not simply describe plans. They also perform the task decomposition required by planners and maintain robustness to changes in objectives~\cite{kannan2024smartllm}.
The integration of perception further tightens this loop, allowing the entire process from observation and orientation to decision and low--level action control~\cite{zager2024towards}. By inserting visual observations and concise scene summaries into prompts, the system can weigh various scenarios, such as obstacle layout and intent prediction, when selecting the next action. This fusion of visual priors with language--driven reasoning equips agents with intelligent capabilities like navigation~\cite{song2025vlm_social_nav}, exploration~\cite{dorbala2024cat_mug} and manipulation learning~\cite{yin2023multi}, demonstrating the breadth of LLM applications in autonomous systems.

Our research follows this trajectory, with the aim of achieving a fundamental shift in the concept of autonomy in confrontations. The model no longer serves as a semantic interpreter, but evolves into a central commander, integrating symbolic reasoning, situational awareness, and tactical planning into a unified architecture.

\section{Methodology}\label{sec:method}

In this section, we introduce the vision--language model--based commander. The model combines perception and decision modules to form a full chain that imitates the human thought process to achieve autonomous command.

\subsection{Confrontation Setup and Preliminaries}
\label{subsec:problem_formulation}

\begin{figure}[htbp]
    \centering
    \includegraphics[width=0.95\textwidth]{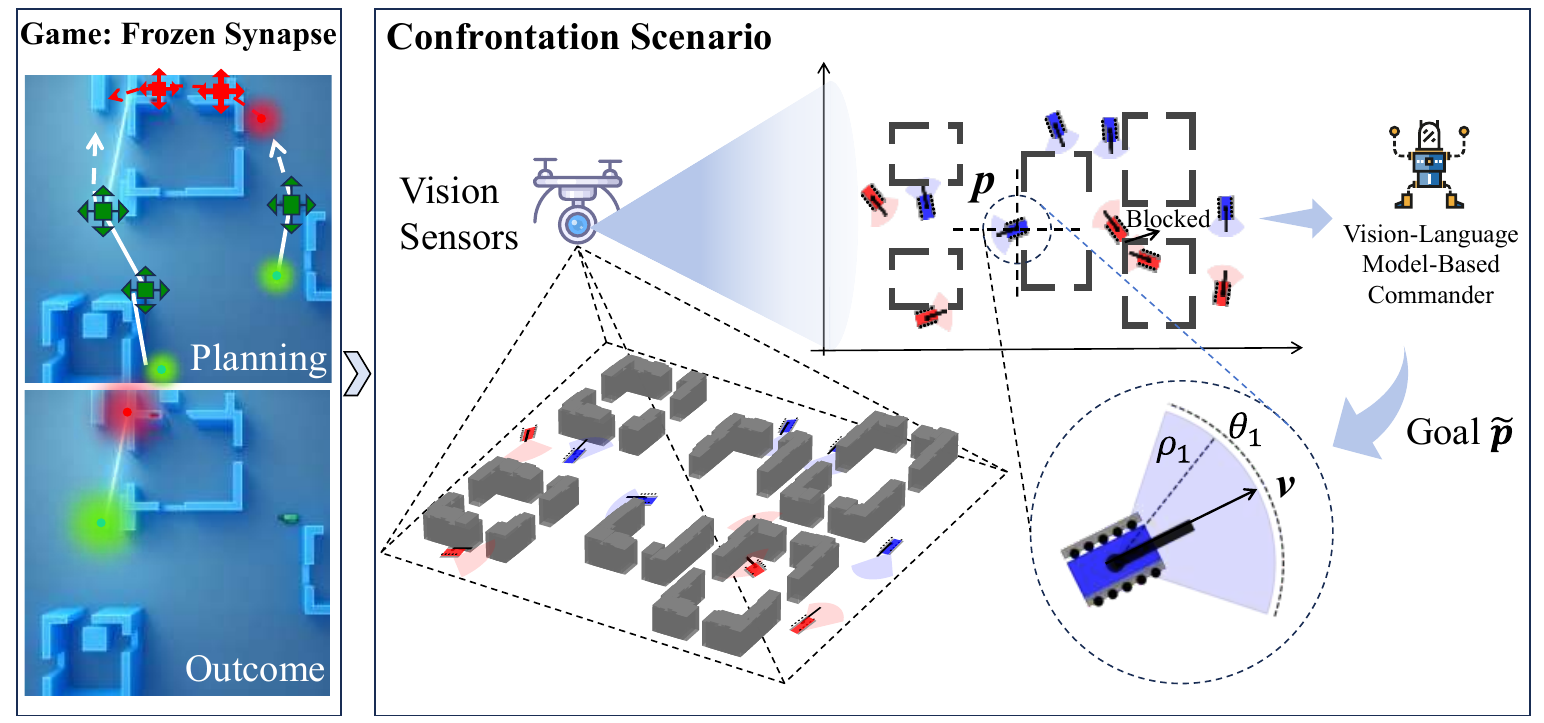}
    \caption{Diagram of the confrontation scenario.}
    \label{fig:scene}
\end{figure}

We consider a ground confrontation scenario in which a top--mounted visual sensor continuously captures the entire arena and converts it into a two--dimensional map, as shown in Figure~\ref{fig:scene}. The environment contains static obstacles that restrict both movement and line of sight. Two symmetric teams, each composed of $N$ autonomous agents, compete in the arena. These teams are denoted as the allied force $\mathbb{A}=\{a_1,\dots,a_N\}$ and the enemy force $\mathbb{E}=\{e_1,\dots,e_N\}$. All agents follow a common set of engagement rules under the same capability constraints. Once an agent is hit, it is considered eliminated and immediately becomes a static obstacle at its final position, creating a dynamic environment. The confrontation ends either when one team is completely eliminated or when a predefined time limit is reached.

At each time step $t$, the navigable region of the environment is indicated by $\Omega_{\mathrm{free}}(t) \subset \mathbb{R}^2$, representing the subset of the plane that is free of obstacles. Each allied agent $a_i$ is represented by its position $\boldsymbol{p}_{a_i}(t)\in\Omega_{\mathrm{free}}(t)$ and velocity $\boldsymbol{v}_{a_i}(t)\in\mathbb{R}^{2}$ with $\lVert\boldsymbol{v}_{a_i}(t)\rVert_{2}\le v_{\max}$, where $\lVert \cdot \rVert_2$ denotes the Euclidean norm. And each agent engages within a limited attack radius $\rho_1$ and a restricted field of view angle $\theta_1$. The model takes a bird's--eye view image of the entire battlefield as input. The output is a set of scheduling instructions $\boldsymbol{U}(t) = \{\boldsymbol{u}_{a_1}(t), \dots, \boldsymbol{u}_{a_N}(t)\}$ that assign a specific command $\boldsymbol{u}_{a_i}(t)$ to each ally containing the next target waypoint.

\subsection{Vision--Language Model--Based Commander}
\label{subsec:commander}
\begin{figure}[htbp]
    \centering
    \includegraphics[width=0.95\textwidth]{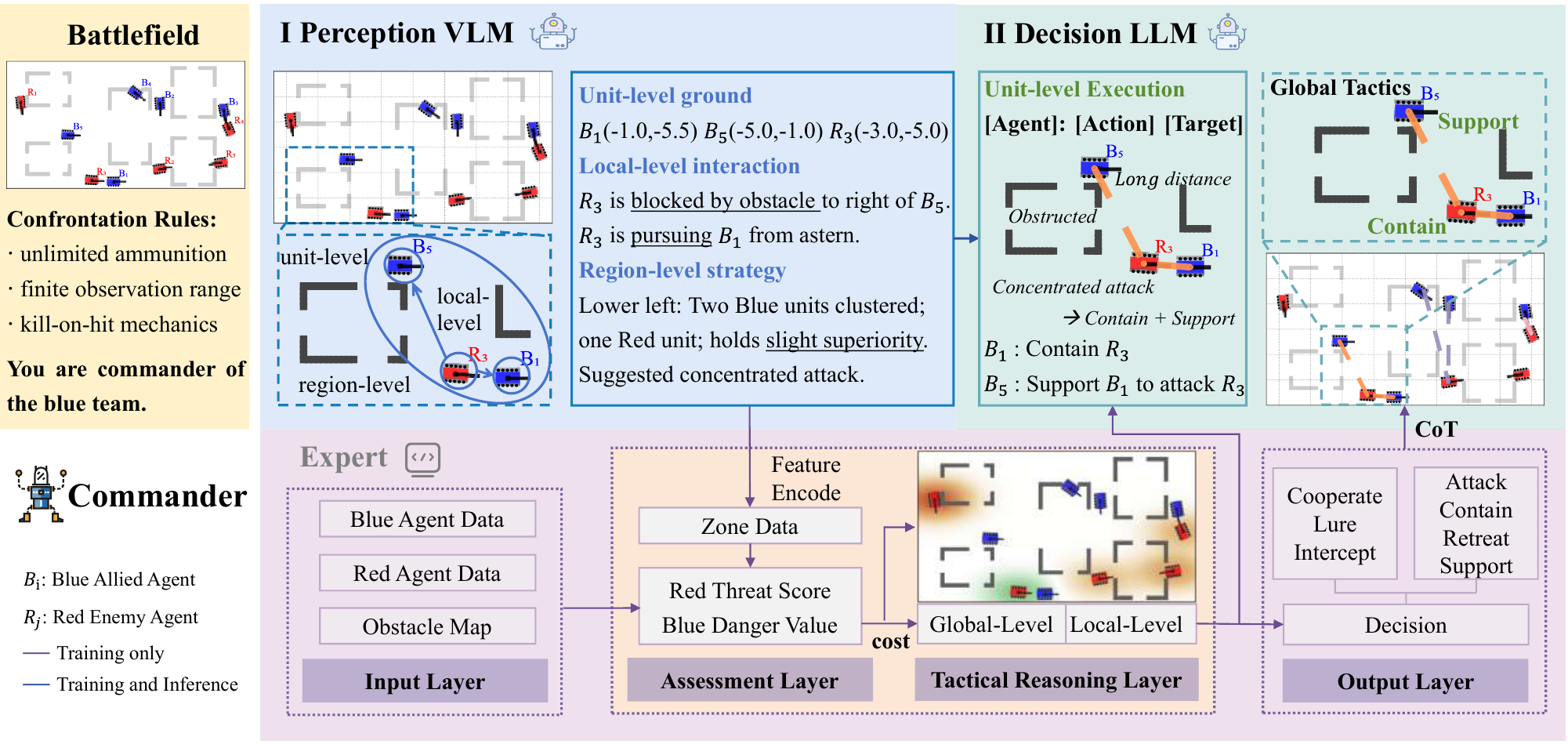}
    \caption{Overview of vision--language model--based commander for tactical decision.}
    \label{fig:overview}
\end{figure}

Figure~\ref{fig:overview} shows the core architecture and details of our model, which consists of three related parts, including the perception module, the decision module, and the expert system used only during training.

As the first part of the model, the perception module is responsible for converting high--dimensional visual input $\mathbf{I} \in \mathbb{R}^{H \times W \times C}$ into semantic representations~$\boldsymbol{S}$ of the environment. Here, \(H\), \(W\), and \(C\) denote the height, width, and number of channels of the input images, respectively. This functionality is achieved through VLM that effectively acts as an intelligent sensor, ensuring that both fine--grained individual behavior and coarse--grained strategic intent are grounded in a consistent perceptual framework.

\begin{equation}
\mathcal{V}: \mathbf{I} \mapsto \boldsymbol{S} = \{ S_u, S_l, S_r \},
\label{eq:vlm_mapping}
\end{equation}
\noindent
where \(\mathcal{V}\) denotes the vision--language model that performs joint visual recognition and semantic abstraction. The output semantic space $\boldsymbol{S}$ is organized into three hierarchical levels of abstraction. $S_u$ represents unit--level grounding, $S_l$ represents local--level interaction, and $S_r$ represents region--level summaries.

For hierarchical output, the detailed content includes,
(i)~\textbf{Unit--level grounding} captures the state of individual agents, including spatial coordinates, team affiliation, and current status. This forms a foundational map of all participating units on the battlefield, providing the basis for situational awareness.
(ii)~\textbf{Local--level interaction} identifies spatial relationships among nearby agents, including visibility conditions, enemy distribution, and environmental constraints. This encodes short--range action pressure. 
(iii)~\textbf{Region--level summary}, constrained by the feasible domain of agent behavior, divides the battlefield into discrete regions. For each region, the model provides a situational summary that includes the distribution patterns of both teams, as well as a qualitative assessment of the regional state.

Through the multiscale abstraction, the perception module generates rich representations that capture both spatial structure and overall situation, enabling the decision module to reason with semantic representations without the need of post--processing.

The second stage of our model is the core reasoning engine. Implemented through the LLM, this module plays the role of a global tactical planner. The LLM works entirely in semantic space, and the structured semantic interface enables the LLM to synthesize scattered perception fragments into a unified battlefield view. Unlike reactive controllers that respond locally, the LLM evaluates urgency and priorities and then allocates strategic goals into unit--level tasks. 
\begin{equation}
\mathcal{L}: \boldsymbol{S}(t) \mapsto \boldsymbol{U}(t) = \left\{ u_{a_1}(t), \, u_{a_2}(t), \dots, u_{a_N}(t) \right\},
\label{eq:llm_instruction_set}
\end{equation}
\noindent
where $\mathcal{L}$ denotes the tactical scheduling function implemented by the LLM, $\boldsymbol{S}(t)$ is the structured semantic input at time $t$. Each instruction~$u_{a_i}(t)$ assigned to agent~$a_i$ is defined as follows.
\begin{equation}
u_{a_i}(t) = \left( a_i, \tau_i, \tilde{\mathbf{p}}_i \right).
\end{equation}
\noindent
where $\tau_i \in \mathcal{T}$ specifies the action type to be executed, and $\tilde{\mathbf{p}}_i \in \mathbb{R}^2$ is the next waypoint. The action type set $\mathcal{T}$ includes
\[
\mathcal{T} = \{Attack, Support, Retreat, Intercept, Lure, Cooperate, \dots \}
\]

This representation method enables scalable and interpretable decision--making. Within the team, LLM resolves role conflicts and balances risks between different agents, ensuring consistency across all tasks. In global planning, it expresses complex tactics through a unified language protocol, such as coordinated attacks, sequential containment, and cross-regional reinforcements. Furthermore, the decision LLM supports combinatorial tactical reasoning, generating higher--order strategies like luring enemies into ambushes.

\subsection{Expert System for Model Training}\label{subsec:expert_system}

To achieve semantic alignment between VLM and LLM and train the model to adapt to confrontations, we introduce a simulation expert system that produces high--quality decision labels. Acting as an agent for human reasoning, the system converts structured perceptual output into consistent tactical instructions. A set of heuristic and rule--based algorithms guide its decision--making logic.

Initially, the expert system systematically processes the perceptual data by computing two essential metrics, namely the threat score for enemies and the danger value for allies. 

The threat score $T(e_j)$ quantifies the potential strategic impact posed by an enemy agent $e_j$. Specifically, it aggregates spatial position, regional weight, and engagement force ratio into a single normalized score.
%--------------------------------------------------------------
% Threat‑score definition
\begin{equation}\label{eq:threat_score}
T(e_j)=
\frac{I\bigl(\boldsymbol{p}_{e_j}\bigr)}
     {\displaystyle \max_{\boldsymbol{x}} I(\boldsymbol{x})},
\qquad
I(\boldsymbol{x})=
\sum_{e_l\in\mathbb{E}}
Z(e_l)\,
M(e_l)\,
\mathrm{e}^{-\|A(\boldsymbol{x}-\boldsymbol{p}_{e_l})\|_2^{2}/(2\sigma^{2})},
\end{equation}

\noindent
where $I(\boldsymbol{x})$ is the intensity field of threat generated by each enemy $e_l\!\in\!\mathbb{E}$. $\max I(\boldsymbol{x})$ represents the maximum value of this field, used as a normalization factor.
$Z(e_l)$ encodes the tactical weight of zone $z_l$.  
$M(e_l)$ is a local force balance modifier (see equation~\eqref{eq:threat_map}). 
$A=\operatorname{diag}(1/\rho,\,1)$ compensates for the aspect ratio of the map $\rho$,
and $\sigma$ is the Gaussian bandwidth.

\begin{equation}\label{eq:threat_map}
M(e_l)=
\frac{|S_{\mathbb{A}}(e_l,r)|+1}
     {|S_{\mathbb{E}}(e_l,r)|+1}\,
\sigma\!\bigl(
        \bar d_{\mathbb{E}}(e_l,r)\big/\bar d_{\mathbb{A}}(e_l,r)
      \bigr),
\end{equation}

\noindent
where $S_{\mathbb{A}}(e_l,r)$ and $S_{\mathbb{E}}(e_l,r)$ are, respectively, the allied and enemy sets within radius $r$ of agent $e_l$. 
$\bar d_{\mathbb{A}}$ and $\bar d_{\mathbb{E}}$ denote the corresponding mean distances.  
$\sigma(\cdot)=\bigl(1+\mathrm{e}^{-x}\bigr)^{-1}$ is the logistic function.
%--------------------------------------------------------------

In parallel, the system calculates a danger value $D(a_i)$ to quantify the exposure of each allied agent $a_i$ to surrounding threats, taking into account visibility, relative position, and the local force distribution of both sides.
%--------------------------------------------------------------
% Danger‑value definition
\begin{equation}\label{eq:danger_value}
\begin{aligned}
D(a_i)=&
\;w_{\text{en}}\lvert S_{\mathbb{E}}(a_i,r)\rvert
-
w_{\text{al}}\lvert S_{\mathbb{A}}(a_i,r)\rvert \\[2pt]
&+\sum_{e_j\in S_{\mathbb{E}}(a_i,r)}
      \frac{w_{\text{d,en}}}{d(a_i,e_j)+\varepsilon}\,
      \alpha_{e_j}^{\text{vis}}\,
      \alpha_{e_j}^{\text{p}}\,
      \alpha_{e_j}^{\text{v}} \\[4pt]
&-\sum_{a_k\in S_{\mathbb{A}}(a_i,r)}
      \frac{w_{\text{d,al}}}{d(a_i,a_k)+\varepsilon}\,
      \beta_{a_k}^{\text{vis}}\,
      \beta_{a_k}^{\text{p}},
\end{aligned}
\end{equation}

\noindent
where the scalar weights $w_{\text{en}}$, $w_{\text{al}}$, $w_{\text{d,en}}$, and $w_{\text{d,al}}$ tune the relative impact of enemy counts, ally counts, and distance--based effects. The factors $\alpha_{e_j}^{\text{vis}}$, $\alpha_{e_j}^{\text{p}}$, and $\alpha_{e_j}^{\text{v}}$ capture, respectively, the visibility, the relative position, and the relative velocity of the enemy $e_j$. Similarly, $\beta_{a_k}^{\text{vis}}$ and $\beta_{a_k}^{\text{p}}$ quantify the degree of protection of an allied agent $a_k$. Higher danger values indicate greater immediate risks to the agent.
%--------------------------------------------------------------

Based on these calculated values, the expert system further calculates the attack cost $C(a_i,e_j)$, which is used to reflect the risks of engaging the enemy agent $e_j$ and to assist in determining whether the tactic is feasible. The attack cost is based on the enemy threat score and basic engagement factors and is calculated as follows.
%--------------------------------------------------------------
% Attack–cost 
\begin{equation}\label{eq:attack_cost_single}
\begin{aligned}
C\!\bigl(a_i,e_j\bigr)=\;
&\;\omega_{\mathrm{d}}\,
   \min\!\bigl(
         \lVert\boldsymbol{p}_{a_i}-\boldsymbol{p}_{e_j}\rVert_2/d_{\max},
         \,1
       \bigr) \\[2pt]
&+\;\omega_{\mathrm{o}}\,
   \frac{1}{\pi}\,
   \arccos\!\Bigl(
     \frac{\boldsymbol{v}_{a_i}}
          {\lVert\boldsymbol{v}_{a_i}\rVert_2}\cdot
     \frac{\boldsymbol{p}_{e_j}-\boldsymbol{p}_{a_i}}
          {\lVert\boldsymbol{p}_{e_j}-\boldsymbol{p}_{a_i}\rVert_2}
   \Bigr) \\[2pt]
&+\;\omega_{\mathrm{t}}\,
     T(e_j)
   -\omega_{\mathrm{vis}}\,
     v(a_i,e_j),  
\end{aligned}
\end{equation}

For cooperative engagements involving multiple allied agents, the cost is modified as follows.
\begin{equation}\label{eq:attack_cost_coop}
\begin{aligned}
C\bigl(a_i,a_k,e_j\bigr)=\,
& C\bigl(a_i,e_j\bigr)
 + C\bigl(a_k,e_j\bigr)  \\[2pt]
&+\gamma\,
  \frac{1}{\pi}\,
  \bigl\lvert
        \arccos\!\bigl(
          \hat{\boldsymbol{u}}_{i j}\cdot\hat{\boldsymbol{u}}_{k j}
        \bigr)-\tfrac{\pi}{2}
  \bigr\rvert  \\[2pt]
&+\zeta\,
  \bigl[Z(a_i)+Z(a_k)\bigr]. 
\end{aligned}
\end{equation}

\noindent
where, $d_{\max}$ caps the distance term. The binary indicator $v(a_i,e_j)\!\in\!\{0,1\}$ indicates the line of sight. $\hat{\boldsymbol{u}}_{x j}=(\boldsymbol{p}_{e_j}-\boldsymbol{p}_{a_x})/ \lVert\boldsymbol{p}_{e_j}-\boldsymbol{p}_{a_x}\rVert_2$. And $\omega_{\mathrm{d}}$, $\omega_{\mathrm{o}}$, $\omega_{\mathrm{t}}$, $\omega_{\mathrm{vis}}$, $\gamma$ and $\zeta$ are positive tuning weights. The higher the value of $C\!\bigl(a_i,e_j\bigr)$ and $C\bigl(a_i,a_k,e_j\bigr)$, the greater the risk of engagement.

% ==========  tab:expert_system  ==================
\begin{table}[htbp]
\caption{Tactical Decision Expert System.}
\label{tab:expert_system}
\renewcommand{\arraystretch}{1.5}
\begin{tabular*}{\textwidth}{@{\extracolsep{\fill}}llll}
\toprule
\textbf{Rule} & \textbf{Scope} & \textbf{Condition} & \textbf{Action} \\
\midrule
R1 & Global & $|\mathbb{E}| = 1$ & \texttt{Attack}$(a_i,e_j)$ \\
R2 & Global & $|\mathbb{A}| = 1 \land |\mathbb{E}| > 1$ & \texttt{Retreat} \\
R3 & Local  & $\exists\, e_j \in S_{\mathbb{E}}(a_i,\epsilon_{\text{engage}})$ s.t. $v(a_i,e_j) = 1$ & \texttt{Attack}$(a_i,e_j)$ \\
R4 & Local  & $\exists\, e_j \in S_{\mathbb{E}}(a_i,\epsilon_{\text{threat}})$ s.t. $v(a_i,e_j) = 0$ & \texttt{Retreat} \\
\midrule
R5 & Global & $\exists\,(a_i,a_k) \in \mathbb{A},\; \exists\, e_j \in \mathbb{E}$ s.t. & \texttt{Contain}$(a_i,e_j)$ \\
   &        & $C(a_i,a_k,e_j) < \theta_{\text{cost}},\; v(a_i,e_j) = 0,$ & \texttt{Support}$(a_k,a_i,e_j)$ \\
   &        & $[d(a_i,a_k) > \theta_{\text{dist}} \lor \theta_{ik}^{(j)} > \theta_{\text{sep}}]$ & \\
R6 & Global & $\exists\,(a_i,a_k)\in\mathbb{A},\; \exists\, e_j \in \mathbb{E}$ s.t. & \texttt{Lure}$(a_i,e_j)$ \\
   &        & $C(a_i,a_k,e_j) < \theta_{\text{cost}},\; v(a_i,e_j) = 0,$ & \texttt{Intercept}$(a_k,a_i,e_j)$ \\
   &        & $[d(a_i,a_k) < \theta_{\text{dist}} \land \theta_{ik}^{(j)} < \theta_{\text{sep}}]$ & \\
R7 & Global & $\exists\,(a_i,a_k)\in\mathbb{A},\; \exists\, e_j \in \mathbb{E}$ s.t. & \texttt{Cooperate}$(a_i,a_k,e_j)$ \\
   &        & $C(a_i,a_k,e_j) < \theta_{\text{cost}} \text{ and } D(a_i), D(a_k) < \theta_{\text{coop}}$ & \\
\midrule
R8 & Local  & $D(a_i) < \delta_{\text{attack}},\; T(e_j) < \delta_{\text{threat}},$ & \texttt{Attack}$(a_i,e_j)$ \\
   &        & and $\exists\, e_j$ s.t. $C(a_i,e_j) < \delta_{\text{cost}}$ & \\
R9 & Local  & $D(a_i) < \delta_{\text{contain}}$ and $\exists\, e_j$ s.t. $C(a_i,e_j) < \delta_{\text{cost}}$ & \texttt{Contain}$(a_i,e_j)$ \\
R10& Local  & $D(a_i) < \delta_{\text{contain}}$ and $\nexists\, e_j$ s.t. $C(a_i,e_j) < \delta_{\text{cost}}$ & \texttt{Support}$(a_i,a_k)$ \\
R11& Local  & $\nexists\, e_j$ s.t. attack / contain / support & \texttt{Retreat} \\
\bottomrule
\end{tabular*}
\footnotetext[1]{Rules R1-R4 are first priority, R5-R7 are second priority, and R8-R11 are third priority.}
\footnotetext[2]{$\theta$,~$\delta$, and~$\epsilon$ are all set thresholds.}
\end{table}

These computational metrics are systematically evaluated through a set of predefined rules, with the specific rules detailed in Table~\ref{tab:expert_system}. The rules are divided into two types, namely global rules and local rules. Global rules are used to handle strategic--level decisions based on the global context, and local rules guide unit--level actions based on the immediate context of an individual agent. The rule sets collectively form a transparent and context--adaptive action framework consistent with expert reasoning, thereby establishing a comprehensive decision--making system to guide LLM.

\subsection{Fine--Tune VLM and LLM}\label{subsec:fine_tuning}

We adopt the pretrained QWEN2.5--VL--7B--Instruct model as the vision language backbone, and the more lightweight QWEN2.5--3B--Instruct model as the base for the natural language inference model, subjecting both to two successive adaptation rounds. 

\begin{figure}[htbp]
    \centering
    \includegraphics[width=0.55\textwidth]{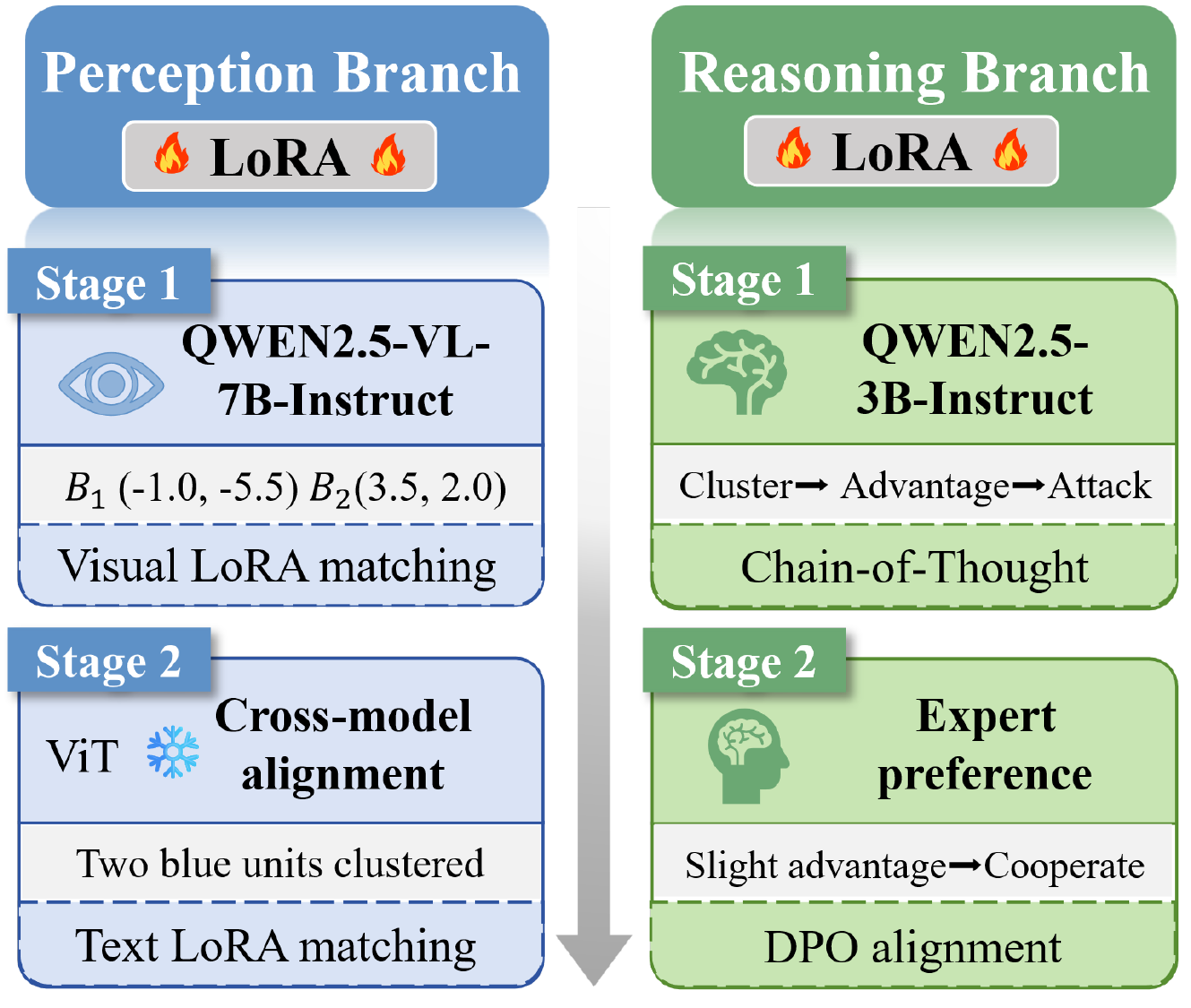}
    \caption{Model training pipeline.}
    \label{fig:fine_tune}
\end{figure}

In the perception branch, the first low--rank adaptation~(LoRA) stage keeps the visual encoder trainable, enabling it to internalize the domain--specific scene layouts and enhance recognition and localization capabilities. From the generated checkpoints, the second stage freezes the visual layers and exposes the remaining cross--modal stack to higher--order courses that integrate situational awareness, contextual comprehension, and structured semantic generation. 

The language branch follows a similar scheme. An initial supervised LoRA fine--tuning emphasizes the CoT exemplars, imprinting explicit reasoning trajectories, after which direct preference optimization (DPO) aligns the planner with the expert strategic priors formalized in Section~\ref{subsec:expert_system}.  

During the fine--tuning process, we used the following adversarial optimization targets. The survival rate~\(R_S\) is defined as the average proportion of allies that survive at the end of the combat, quantifying the overall survival advantage.
\begin{equation}
  R_S \;=\;
  \frac{\bar S_{\mathrm{b}}}
       {\bar S_{\mathrm{b}}+\bar S_{\mathrm{r}}}
  \;\in\,[0,1],
  \label{eq:survival_ratio}
\end{equation}
where \(\bar S_{\mathrm{b}}\) and \(\bar S_{\mathrm{r}}\) denote the mean number of surviving allies and enemies,  respectively, averaged over \(N\) independent trials.

At the same time, since decisions are not made at every step, we define the decision model driving rate~$R_{drv}$ as the percentage of the total number of actions taken by the allied agent originating from the decision model. In order to quantitatively characterize the gains from decision driving, we propose the decision gain index $I_{\text{drv}}$ as the ratio of the allied survival rate to the LLM drive rate.
\begin{equation}
    I_{\text{drv}} = \frac{R_{\text{S}}} {R_{\text{drv}}}.
\end{equation}

\section{Experiments}\label{sec:experients}

In this section, we conduct extensive comparative experiments with different baseline models to evaluate the strategic performance of our model. We also compare models without expert system training to demonstrate its importance. In addition, we test the generalization of the model in more agent settings by extending the model trained with 5 agents against 5 agents to 7 against 7 and 9 against 9 environments.

\subsection{Experimental Setup}

Before presenting the experiment results, we first detail the configuration of the autonomous confrontation system simulator. All experiments are carried out in a rectangular arena measuring $30\,\mathrm{m}\times16\,\mathrm{m}$ that initially deploys an equal number of allied and enemy UGVs placed uniformly at random. The attack radius is fixed at $\rho_{1}=1\,\mathrm{m}$, and every static obstacle is represented as a square object with dimensions $0.3\,\mathrm{m} \times 0.3\,\mathrm{m}$. The motion is subject to maximum speed restriction $\lVert\boldsymbol{v}\rVert_{\max}=2\,\mathrm{m/s}$ with a discrete step $\delta_t=0.1\,\mathrm{s}$.
The maximum adversarial step size is set to 2000 steps.

During training, we set up five agents for each of the blue and red teams. All agents of both teams use the same motion controller and geometric path--planning algorithm. Each experiment is repeated with independent random seeds to reduce sampling bias. Ten independent experiments are conducted for different combinations of the allied and enemy decision models, and all reported metrics are based on the average values of these experiments.

In addition, we train a decision model consisting solely of VLM, which generates all perception and decision content at once. It undergoes the same supervised fine--tuning and DPO alignment as our model, with the same hyperparameters and cosine learning rates.

\subsection{Comparative Analysis}

We set two fixed enemy strategies and simulated confrontations with different baseline models of allied decision and our method. The baseline models include rule--based model~(denoted as Rule), reinforcement learning--based model~(denoted as RL), and a single visual language model trained by the same expert system~(denoted as VLM).

To evaluate the performance of the perception module and the overall adversarial results, we set perception metrics and confrontation metrics separately. Perception metrics include perception accuracy~\(P\), recall rate~\(R\), and hallucination rate~$R_{H}$. The confrontation metrics include average decision time~$\overline{T}_{c}$, win rate~$R_\mathrm{W}$, survival rate~\(R_S\), and decision gain index~$I_{\text{drv}}$, as defined in Section~\ref{subsec:fine_tuning}. Table \ref{tab:comparative_experiment} reports the average results of all experiments.

\begin{table}[htbp]
\caption{Comparative experimental results of our method and baseline models on different opponent decision strategies.}\label{tab:comparative_experiment}
\begin{tabular*}{\textwidth}{@{\extracolsep\fill}llccccccc}
\toprule
\begin{tabular}[c]{@{}c@{}}Enemy \\ Decision \\ Model\end{tabular} & 
\begin{tabular}[c]{@{}c@{}}Allied \\ Decision \\ Model\end{tabular} & 
\multicolumn{3}{@{}c@{}}{Perception Metrics} & 
\multicolumn{4}{@{}c@{}}{Confrontation Metrics} \\
\cmidrule(lr){3-5} \cmidrule(lr){6-9}
& & $R$ & $P$ & $R_{\mathrm{H}}$ & $\overline{T}_{\mathrm{c}}(s)$ & $R_{\mathrm{W}}(\%)$ & $R_{\mathrm{S}}$ & $I_{\mathrm{drv}}$ \\
\midrule
\multirow{4}{*}{Rule} 
& vs. Rule & - & - & - & - & 58 & 0.56 & 0.56 \\
& vs. RL   & - & - & - & - & 67 & 0.78 & 0.78 \\
& vs. VLM  & 0.67 & 0.61 & 0.33 & 17.8 & 67 & 0.79 & 1.05 \\
& vs. Ours & \textbf{0.76\,$\uparrow$} & 0.68 & 0.24 & \textbf{15.43\,$\downarrow$} & 83 & 0.90 & \textbf{1.3\,$\uparrow$} \\
\midrule
\multirow{4}{*}{RL} 
& vs. Rule & - & - & - & - & 72 & 0.76 & 0.76 \\
& vs. RL   & - & - & - & - & 56 & 0.67 & 0.67 \\
& vs. VLM  & 0.76 & 0.75 & 0.24 & 20.84 & 71 & 0.85 & 1.29 \\
& vs. Ours & 0.79 & 0.79 & 0.21 & \textbf{14.49\,$\downarrow$} & 80 & 0.86 & \textbf{1.32} \\
\bottomrule
\end{tabular*}
\end{table}

The results show that introducing lightweight language reasoning based on visual grounding simultaneously enhances perceptual quality, shortens decision time, and leads to better performance. 

Compared with the single VLM, the perception module of our model achieves higher precision and recall.
Freed from the burden of tactical reasoning and long--text output, the visual backbone can focus on understanding images and generating clearer scene descriptions. At the same time,  we notice that our model reduced the average decision time by approximately 25\%,  and the resulting response acceleration does not reduce the quality of the decision, but rather translated into higher survival rates and decision gain indices.

Therefore, when supplied with accurate situational cues, even a lightweight language reasoning component is sufficient to support complicated reasoning. Our model design achieves a favorable balance between computational cost and adversarial gains.

\begin{figure}[htbp]
    \centering
    \includegraphics[width=0.95\textwidth]{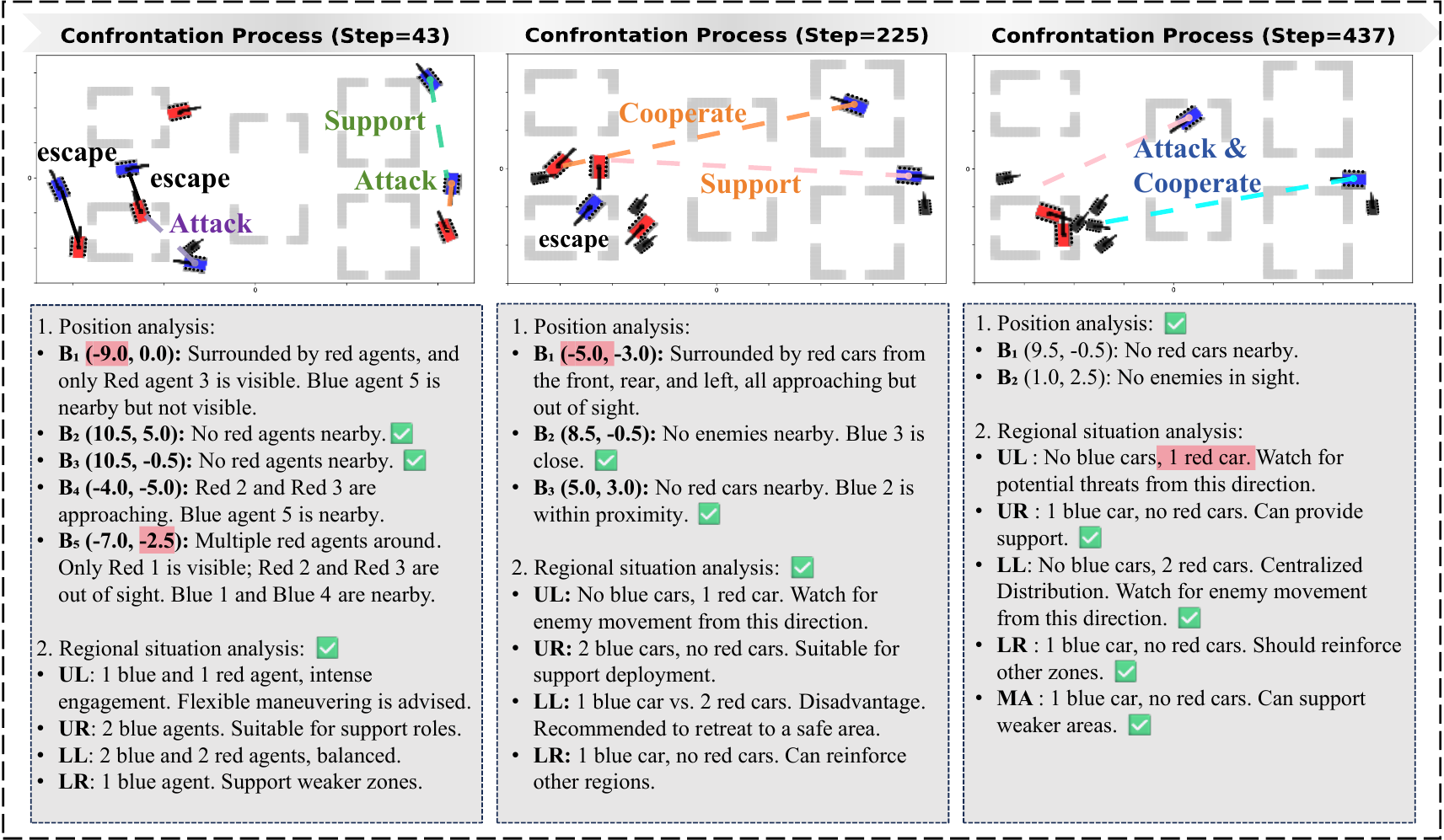}
    \caption{An example of the adversarial process and model output. Content marked in red indicates deviation and incorrect output, and content marked in green indicates completely correct content.}
    \label{fig:confrontation_process}
\end{figure}

\begin{figure}[htbp]
    \centering
    \includegraphics[width=1\textwidth]{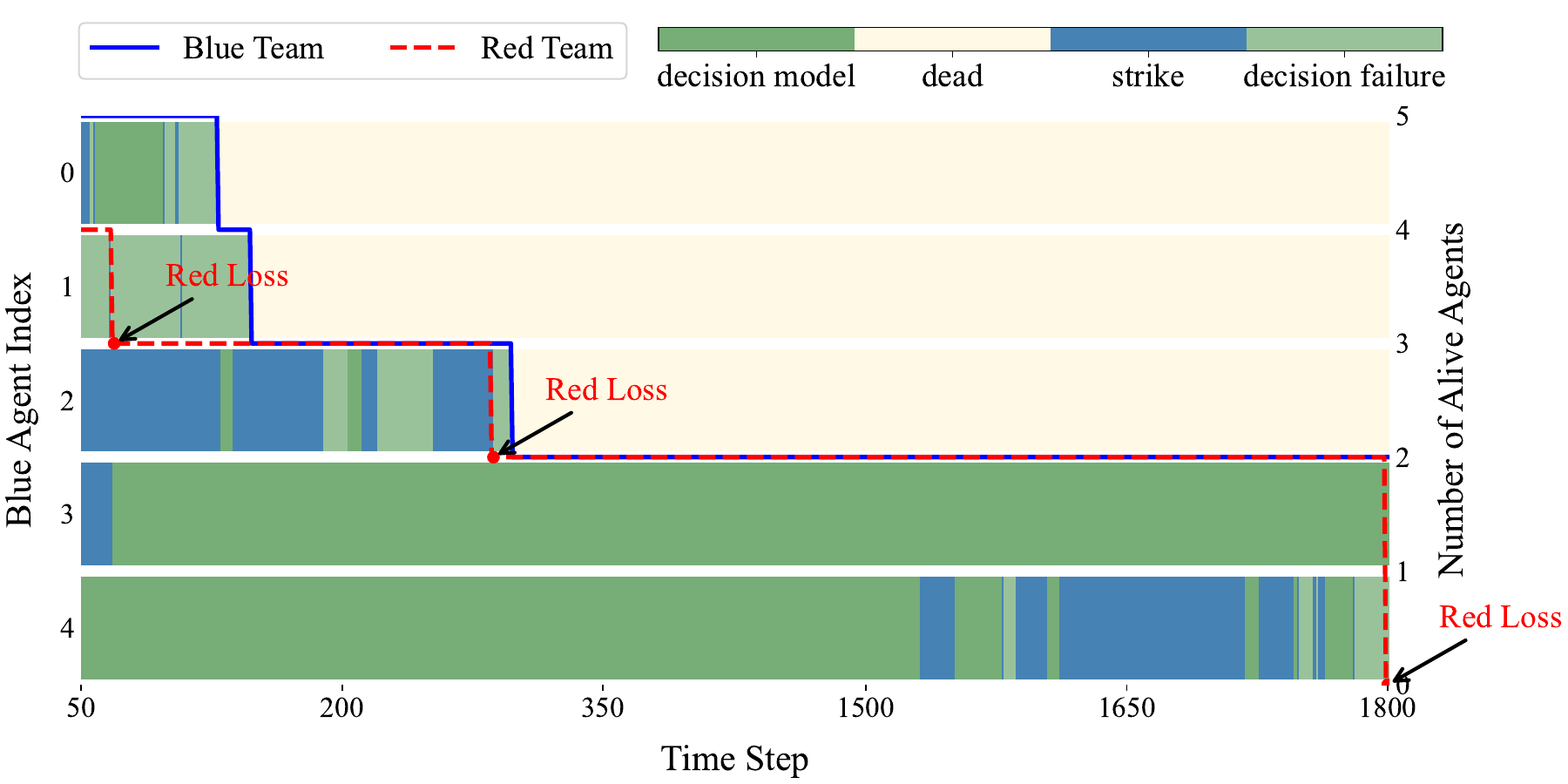}
    \caption{Decision situation of each blue agent during the confrontation process, and the survival curves of both teams corresponding to the example.}
    \label{fig:decision}
\end{figure}

We show a confrontation process in Figure~\ref{fig:confrontation_process}, and Figure~\ref{fig:decision} shows the decisions of each blue agent (i.e., our approach) and the survival curves of both teams during the confrontation. Throughout the episode, VLM perception remains mostly reliable at the semantic level, with agent identities, relative positions, and occlusion states consistently recognized. However, localization errors occur when multiple agents are clustered, especially when positioned behind obstacles. The most obvious deviation appears in the leftmost subplot, where inaccuracies in the coordinates of two blue agents directly affect their immediate scheduling by the LLM.

Despite this, the learned reasoning layer proved to be flexible and both global and local scheduling remain effective. In the rightmost figure, the LLM exploits geographical conditions created by obstacles to assign two blue units separately. This results in an encircling formation against the two red units, rather than a joint strike from the upper exit targeting a single enemy. It is at this moment that our commander exhibits tactical potential beyond those of expert systems. These decisions are further validated in subsequent steps, as shown in Figure~\ref{fig:decision}, where red units are eliminated.

This confrontation case demonstrates that under visually cluttered conditions, perception hallucinations are largely limited to localization precision. On the other hand, the LLM can tolerate such noise and make reasonable decisions relying on geometric relationships and visibility constraints.

\subsection{Ablation and Generalization Experiments}

To quantify the contribution of each module and expert system, we perform ablation experiments under the same environmental conditions. Table~\ref{tab:ablation} compares the performance changes in different configurations.

The improvement in the decision--making system's performance relies on the synergistic integration of multiple modules. Introducing the visual--language model or language model alone is insufficient to effectively address dynamic multi--agent game environments and cannot significantly enhance overall adversarial performance.
Although the perception module maintained comparable perception accuracy, the baseline model without expert training exhibits poorer performance and suffers the longest decision time. The single VLM trained with experts compensates for some shortcomings, but our VLM$+$LLM architecture still achieves the best overall results. The results demonstrate that the high--quality reasoning chain supervision and preference optimization provided by the expert system are important factors in enhancing the stability of the model's tactical reasoning.

\begin{table}[htbp]
\caption{Ablation experiments results.}
\label{tab:ablation}
\centering
\begin{tabular*}{\textwidth}{@{\extracolsep\fill}c ccc cc ccc}
\toprule
\multirow{3}{*}{\begin{tabular}{c}Enemy \\ Decision \\ Model\end{tabular}} 
& \multicolumn{3}{c}{Allied Decision Model} 
& \multirow{2}{*}{$R$} 
& \multirow{2}{*}{$\overline{T}_{\mathrm{c}}(s)$} 
& \multirow{2}{*}{$R_{\mathrm{W}}(\%)$} 
& \multirow{2}{*}{$R_{\mathrm{S}}$} 
& \multirow{2}{*}{$I_{\mathrm{drv}}$} \\
\cmidrule(lr){2-4}
& VLM & LLM & Expert\\
\midrule
\multirow{4}{*}{Rule} 
& $\checkmark$ & $\times$ & $\times$ & 0.57 & 26.4 & 63 & 0.62 & 1.09 \\
& $\checkmark$ & $\checkmark$ & $\times$ & 0.72 & 24.83 & 67 & 0.73 & 1.12 \\
& $\checkmark$ & $\times$ & $\checkmark$ & 0.67 & 17.80 & 67 & 0.79 & 1.05 \\
& $\checkmark$ & $\checkmark$ & $\checkmark$ & \textbf{0.76} & \textbf{15.43} & \textbf{83} & \textbf{0.90} & \textbf{1.30} \\
\midrule
\multirow{4}{*}{RL} 
& $\checkmark$ & $\times$ & $\times$ & 0.56 & 33.48 & 67 & 0.69 & 1.17 \\
& $\checkmark$ & $\checkmark$ & $\times$ & 0.78 & 29.35 & 71 & 0.79 & 1.15 \\
& $\checkmark$ & $\times$ & $\checkmark$ & 0.76 & 20.84 & 71 & 0.85 & 1.29 \\
& $\checkmark$ & $\checkmark$ & $\checkmark$ & \textbf{0.79} & \textbf{14.49} & \textbf{80} & \textbf{0.86} & \textbf{1.32} \\
\bottomrule
\end{tabular*}
\end{table}

\begin{figure}[htbp]
    \centering
    \includegraphics[width=0.95\textwidth]{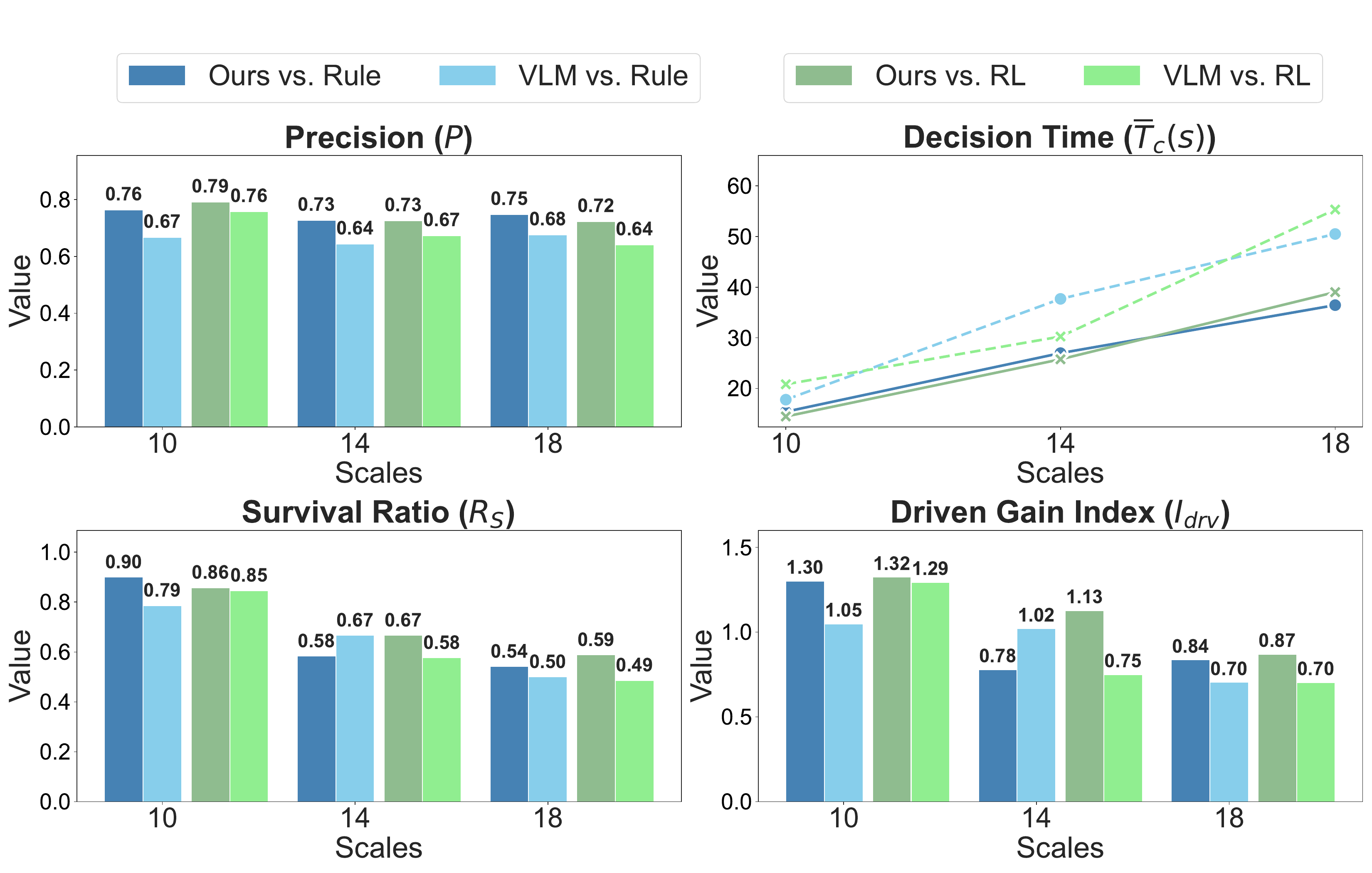}
    \caption{Performance of the proposed method and the VLM only when the scale grows.}
    \label{fig:generalization}
\end{figure}

At the same time, we replicated the baseline adversarial protocol on a larger scale, setting the allied strategies to our model and VLM. Figure~\ref{fig:generalization} reports the results of the comparison of four indicators.

As the density of agents increases, the perception accuracy of a single VLM decreases, while the perceptron in our method always maintains an accuracy of around 75\%. This phenomenon once again confirms the advantage of hierarchical models. It is obvious that the decision time increases as expected with the increase in scale, but the growth rate of the single VLM strategy is particularly sharp compared.

These observations further confirm that separating perception from lightweight semantic planning achieves a balance between computational consumption and decision efficiency, and highlights the robustness of the proposed method in the deployment of large--scale unmanned systems.

\section{Conclusion}\label{sec:conclusions}

In this paper, we have proposed the vision--language model--based commander, which combined visual language perception with a lightweight language planner to address the problem of autonomously generating tactical decisions in confrontations. 
Experimental results have confirmed that by separating visual perception from tactical reasoning, VLM has been able to focus on image understanding to achieve better perceptual results and faster response speed. The addition of a lightweight LLM has improved decision performance, achieving a balance between computational cost and adversarial benefits. Furthermore, this method has enabled robust and scalable decision capabilities, ultimately allowing the commander to develop strategic potential. 
Despite these advancements, we acknowledge that in dense scenarios, the perception module may experience positioning errors and hallucinations, and probably misguide the planner. Therefore, we will focus on enhancing perception to overcome these limitations and extend this semantic--based paradigm to broader applications.

\backmatter

\section*{Acknowledgements}

This work was supported by the National Natural Science Foundation of China under Grant 62088101 and Grant 62003015.

\bibliography{sn-bibliography}% common bib file
%% \input output.bbl

\end{document}